\def\year{2020}\relax
\documentclass[letterpaper]{article} %
\usepackage{amssymb}
\usepackage{bm}
\usepackage{amsmath,amsthm}
\usepackage{algorithmic}
\usepackage{algorithm}
\usepackage{aaai20}  %
\usepackage{times}  %
\usepackage{helvet} %
\usepackage{courier}  %
\usepackage[hyphens]{url}  %
\usepackage{graphicx} %
\usepackage{graphicx}  %
\newcommand{\citet}[1]{\citeauthor{#1}~\shortcite{#1}}
\newcommand{\citep}{\cite}

\def\jl#1{\bigskip{\bf #1}\bgroup\it\ }
\def\ejl{\egroup\par\bigskip}
\newtheorem{theorem}{Theorem}

\newtheorem{lemma}{Lemma}
\newtheorem{prop}{Proposition}

\newtheorem{example}{Example}
\newcommand{\bea}{\begin{eqnarray}}
\newcommand{\eea}{\end{eqnarray}}
\newcommand{\Bea}{\begin{eqnarray*}}
	\newcommand{\Eea}{\end{eqnarray*}}
\newcommand{\ba}{\begin{array}}
	\newcommand{\ea}{\end{array}}
\newcommand{\bt}{\begin{tabular}}
	\newcommand{\et}{\end{tabular}}
\newcommand{\btb}{\begin{table}}
	\newcommand{\etb}{\end{table}}
\newcommand{\bc}{\begin{center}}
	\newcommand{\ec}{\end{center}}
\newcommand{\beq}{\begin{equation}}
\newcommand{\eeq}{\end{equation}}
\newcommand{\E}{\mathbb E}

\newcommand{\cH}{\mathcal{H}}

\newcommand{\cF}{\mathcal F}
\newcommand{\cX}{\mathcal X}
\newcommand{\cY}{\mathcal Y}
\newcommand{\cM}{\mathcal M}

\newcommand{\bP}{\mathbb P}
\newcommand{\bR}{\mathbb R}

\newcommand{\KL}{\mathrm{KL}}

\title{Meta-Learning PAC-Bayes Priors in Model Averaging}
\author{Yimin Huang\\[0.05in]
Huawei Noah's Ark Lab\\[0.05in]
{yimin.huang@huawei.com}%
\And
Weiran Huang\thanks{Correspondence to Weiran Huang.}\\[0.05in]
Huawei Noah's Ark Lab\\[0.05in]
{weiran.huang@outlook.com}%
\And
Liang Li\\[0.05in]
Huawei Noah's Ark Lab\\[0.05in]
{liliang103@huawei.com}%
\And
Zhenguo Li\\[0.05in]
Huawei Noah's Ark Lab\\[0.05in]
{li.zhenguo@huawei.com}}
\usepackage{xcolor}
\usepackage{hyperref}
\usepackage{url}

\definecolor{citecolor}{HTML}{0071BC}
\definecolor{linkcolor}{HTML}{ED1C24}
\hypersetup{colorlinks=true, linkcolor=linkcolor, citecolor=citecolor,urlcolor=black}

\begin{document}
	
	\maketitle

	\begin{abstract}
		Nowadays model uncertainty has become one of the most important problems in both academia and industry. In this paper, we mainly consider the scenario in which we have a common model set used for model averaging instead of selecting a single final model via a model selection procedure to account for this model's uncertainty to improve the reliability and accuracy of inferences. Here one main challenge is to learn the prior over the model set. To tackle this problem, we propose two data-based algorithms to get proper priors for model averaging. 
		One is for meta-learner, the analysts should use historical similar tasks to extract the information about the prior. The other one is for base-learner, a subsampling method is used to deal with the data step by step. 
		Theoretically, an upper bound of risk for our algorithm is presented to guarantee the performance of the worst situation.
		In practice, both methods perform well in simulations and real data studies, especially with poor-quality data.
	\end{abstract}
	
	\section{Introduction}
	It is very common in practice that the distributions generating the observed data are described more adequately by multiple models. A standard procedure to make the inference is that according to some criteria, such as model predictive ability, model fitting ability, and many different information criteria %
	the best model is chosen and assumed as the true model. After selection, all the inferences and conclusions are made based on the assumption.
	
	However, the drawbacks of this approach exist. The selection of one particular model may lead to riskier decisions since it ignores the model uncertainty. In other words, if we choose the wrong model, the consequence will be disastrous. \citet{Moral2015Model} already pointed out the concern, ``From a purely empirical viewpoint, model uncertainty represents a concern because estimates may well depend on the particular model considered.'' Therefore, combining multiple models to reduce the model uncertainty is very desirable.
	
	As an alternative strategy, combining multiple models which is called model averaging enables researchers to draw conclusions based on the whole universe of candidate models. In particular, researchers estimate all the candidate models and then compute a weighted average of all the estimates.
	There are two different approaches to model averaging in the literature, including Frequentist Model Averaging (FMA) and Bayesian Model Averaging (BMA).
	Frequentist approaches focus on improving prediction and use the weighted mean of estimates from different models while Bayesian approaches focus on the probability that a model is true and consider priors and posteriors for different models. 
	
	The FMA approach does not consider priors, so the corresponding estimators depend solely on data. For its simplicity, the FMA approach has received some attention over the last decade. See \citet{Yang2001Adaptive}, \citet{leung2006information}, \citet{hansen2007least} and a detailed review \citet{wang2009frequentist} for reference. 
	
	\citet{leamer1978specification} suggested to use Bayesian inference to reduce the model uncertainty as a framework and pointed out the importance of the fragility of regression analysis to arbitrary decisions about the choice of control variables. Bayesian Model Averaging considers model uncertainty through the prior distribution. The model posteriors are obtained by Bayes' theorem, and therefore allow for combined estimation and prediction.
	Compared with the FMA approaches, there is a huge literature on the use of BMA in statistics.
	
	Influenced by \citet{leamer1978specification}, most works were concentrated on the linear models only. \citet{Raftery1996Approximate} extended in generalized linear models by providing a straightforward approximation. For more details, refer to a landmark review \citet{Moral2015Model} on BMA.
	
	The Bayesian approaches have the advantage of using arbitrary domain knowledge through a proper prior. However, as commented by \citet{hjort2003frequentist}, how to set prior probabilities and how to deal with the priors when they conflict with each other are still problems. The PAC-Bayes framework, first formulated by \citet{Mcallester1999PAC}, was proposed to take the priors into account. 
	In the beginning, most works assumed that loss functions were bounded. For detailed information, see \citet{catoni2007pac}. For unbounded loss, \citet{catoni2004statistical} provided a result under exponential moment assumptions. In the last decade, it has been widely developed. Different types of PAC-Bayes bounds were presented under various assumptions, for example, \citet{Seeger2002PAC}, \citet{seldin2010pac}, \citet{guedj2013pac}, \citet{alquier2016properties}, \citet{grunwald2016fast}, \citet{catoni2016pac}, \citet{lugosi2019regularization}, and \citet{alquier2018simpler}. And, many distribution-dependent priors are used to derive tighter PAC-Bayes bounds such as
	\citet{Lever2013Tighter}, \citet{oneto2016pac}, \citet{dziugaite2018data} and \citet{rivasplata2018pac}. Here, we must distinguish between obtaining the tighter bounds by distribution-dependent priors and using part of the data to meta-learn a prior and the rest to learn the function.
	
	Note that for getting the posterior distribution of the weights, \citet{Ye2016Sparsity} gave a method without choosing a proper prior. For meta-learning the prior, some meta-learners \cite{finn2017model,li2017meta,amit2018meta-learning} are limited to their use of gradient. \citet{amit2018meta-learning} provided an extended PAC-Bayes bound for learning the proper priors. Nevertheless, it involved reusing the data which increased the probability of overfitting, and they gave an implementation only for the normal distribution of the weights.

	In this paper, we propose a specific risk bound under our settings and two data-based methods for adjusting the priors in the PAC-Bayes framework. And, two practical algorithms are given accordingly. The main contributions of this work are the following.
	First, when the historical data existed, we use similar old tasks to extract mutual knowledge with the current task for adjusting the priors. 
	Second, a sequential batch sampling method is proposed to deal with the base-learner for learning posterior by subsampling with the rules made by researchers.
	Third, two theoretical risk bounds are provided for these two situations respectively. Fourth, empirical demonstration shows that the proposed meta-methods have excellent performances in numerical studies.

	The remainder of this paper is organized as follows. In Section \ref{sec:meta}, an upper bound for the averaging model and a practical historical data related algorithm are established for obtaining a better prior. In case that there is no historical data, Section \ref{sec:base} proposes another method called a sequential batch sampling algorithm to adjust the prior step by step. Illustrative simulations including regression and classification tasks given in Section \ref{sec:sim} show that our algorithms will lead to more effective prediction. We further apply the proposed methods to two real datasets and confirm the higher prediction accuracy of the minimizing risk bound method. 
	Some proofs of theories are delegated to the supplementary materials.

	\section{Learning the Prior in Meta-Learner}\label{sec:meta}
	
	In a traditional supervised learning task, the learner needs to find an optimal {\em model} (or hypothesis) to fit the data and then uses the learned model to make predictions.
	In the Bayesian approach, various models are allowed to fit the data. In particular, the learner needs to learn an optimal model {\em distribution} over the candidate models and then uses the learned model distribution to make predictions.

	More specifically, in a supervised learning task, we are given a set $S=\{(x_i, y_i)\}_{i=1}^n$ of i.i.d.\ samples drawn from an unknown distribution $D$ over $\cX\times\cY$, i.e., $(x_i,y_i)\sim D$.
	The goal is to find a model $h$ in the candidate model set $\cH$, a set of functions mapping features (feature vector) to responses, that minimizes the expected loss function $\E_{(x, y)\sim D}L(h, x, y)$, where $L$ is a bounded loss function.
	Without loss of generality, we assume $L$ is bounded by $[0,1]$. %
	Note that this assumption is often used at the beginning of the PAC-Bayes framework. This paper uses McAllester's bound, so this assumption is necessary. Note that our procedure can use other, possibly tighter bounds. In other words, under different regularization conditions, it can be replaced by many other PAC-type bounds as long as the assumption matches the corresponding PAC bound. 
	In the Bayesian framework, a distribution $Q$ over $\cH$ is the purpose instead of searching a specific optimal model $h\in \cH$.
	Therefore, the goal turns to find the optimal model distribution $Q$, which minimizes $\E_{h\sim Q}\E_{(x, y)\sim D}L(h, x, y)$.
	Then one could use the weighted average of these models over $\cH$ to make predictions, namely, $\hat y=\E_{h\sim Q}h(x)$.
	More generally, we further assume that the candidate model set $\cH$ consists of $K$ classes of models $\cM_1, \cM_2, \dots, \cM_K$ with %
	$\cH=\bigcup_{k=1}^K\cM_k$.
	Each model class $\cM_k$ is associated with a probability $w_k$, and
	for each model class $\cM_k$, there is a distribution $Q_k$ over $\cM_k$.
	For example, a model class $\cM_k$ could be a group of models obtained from the Lasso method, and the hyper-parameter $\lambda$ in Lasso follows a distribution $Q_k$.
	Another common example is that $\cM_k$ is a group of neural networks with a certain architecture, and the hyperparameters of neural networks follow a joint distribution $Q_k$.
	In this way,  the total distribution over $\cH$ can be written as
	$\xi=(\bm w, Q_1, \dots, Q_K)$,
	where $\bm w$ consists of $w_1, \dots, w_K$ with $||\bm w||_1=1$.
	The goal of the learning task is to find an optimal distribution $\xi$, the posterior of $h$, which minimizes the expected risk $R(\xi,D):=\E_{h\sim \xi}\E_{(x, y)\sim D}L(h, x, y)$, and then the prediction is made by $\hat y=\E_{h\sim \xi}h(x)=\sum_{k=1}^K [w_k \cdot\E_{h\sim Q_k}h(x)]$.
	
	Since sample distribution $D$ is unknown, the expected risk $R(\xi,D)$ cannot be computed directly.
	Therefore, it is usually approximated by the empirical risk $\hat{R}(\xi,S):=\E_{h\sim \xi}\sum_{(x_i,y_i)\in S} L(h,x_i,y_i)/|S|$ in practice, and $\xi$ is learned by minimizing the empirical risk $\hat{R}(\xi,S)$.
	When the sample size is large enough, it would be a good approximation.
	However, in many situations, we do not have so much data, which may lead to a large difference between them.
	Thus, using the empirical risk $\hat{R}(\xi,S)$ to approximate the expected risk $R(\xi,D)$ is not appropriate any longer.
	
	We first study the difference between the empirical risk $\hat{R}(\xi,S)$ and the expected risk $R(\xi,D)$.
	Based on the literature \cite{Mcallester1999PAC}, we can obtain an upper bound of their difference which is stated as the following lemma.
	
	\begin{lemma}\label{thm:single}
		Let $\xi^0$ be a prior distribution over $\cH$ that must be chosen before observing the samples, and let $\delta\in(0,1)$. Then with probability at least $1-\delta$, the following inequality holds for all posterior distributions $\xi$ over $\cH$,
		\begin{align}\label{eq:single}
		&R(\xi,D)\leq\hat{R}(\xi,S)\nonumber\\
		&+\sqrt{\frac{\KL(\bm{w}||{\bm w^0})+\sum_{k=1}^K w_k \KL(Q_k||Q_{k}^0)+\ln\frac{n}{\delta}}{2(n-1)}},
		\end{align}
		where $n$ is the cardinality of sample set $S$, and $\KL(\cdot || \cdot)$ is the Kullback-Leibler (KL) divergence between two distributions\footnote{$\KL(P||P^0)$ is defined as $\E_{x\sim P}\ln\frac{P(x)}{P^0(x)}$.}.
	\end{lemma}
	
	According to the above lemma, it reveals two facts. (a) The upper bound can be divided into two terms. The first term is called sample complexity caused by the randomness of sampling. The second term is called model complexity caused by the difference between the true distribution of data generation models and the distribution learned for predicting the data. (b) It is clear that only when the sample size $n$ is large, the difference $R(\xi,D)-\hat{R}(\xi,S)$ can be guaranteed to be small.
	Thus, minimizing $\hat{R}(\xi,S)$ may not lead to the minimizer of $R(\xi,D)$, which matches our intuition.
	To avoid the risk of the approximation, one can minimize the upper bound of the expected risk $R(\xi,D)$ instead of using the empirical risk $\hat{R}(\xi,S)$ as an approximation.
	In particular, we denote the right-hand side of Eq.\eqref{eq:single} by $\overline R(\xi,\xi^0,S)$.
	Then one can learn the model distribution $\xi$ by minimizing $\overline R(\xi,\xi^0,S)$.
	Intuitively, such a choice of $\xi$ for the learning task makes the worst case best.
	
	Lemma~\ref{thm:single} also indicates that the prior $\xi^0$ plays an important role.
	Since the choice of $\xi$ balances the tradeoff between the empirical risk $\hat{R}(\xi,S)$ and the regularization term,
	if the prior $\xi^0$ is far away from the true optimal model distribution $\xi^*$, the posterior $\xi$ will also be bad.
	The best situation for optimizing the posterior $\xi$ is that the prior $\xi^0$ exactly equals the true optimal model distribution $\xi^*$.
	Then, the regularization term disappears.
	In other words, if there is a good prior $\xi^0$ which is close to $\xi^*$, the upper bound $\overline R(\xi,\xi^0,S)$ will be small.
	However, without any prior knowledge, one can only use data to help obtain a better prior.
	The naive method is directly using the non-informative prior as $\xi^0$ for minimizing $\overline R(\xi,\xi^0,S)$ to get the posterior $\xi$.
	
\begin{figure}
		\centering
		\includegraphics[width=8.5cm]{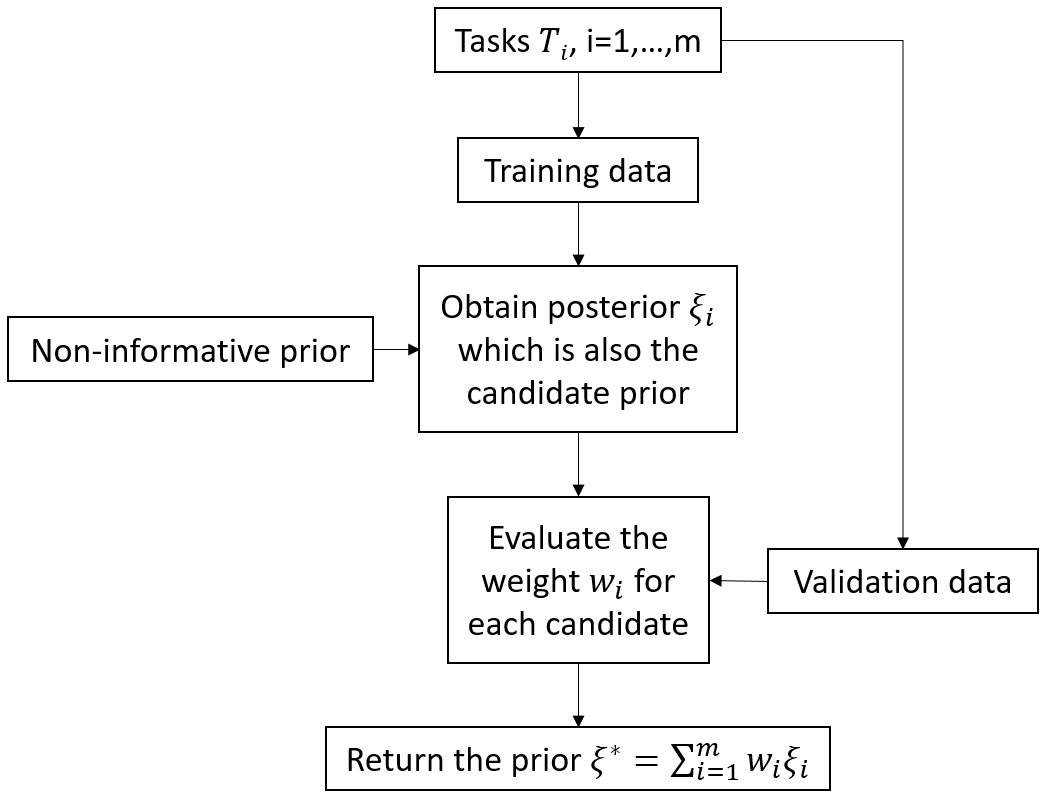}\\
		\caption{Illustration of the meta-learning process.}\label{fig:ill}
	\end{figure}

	When the extra data of historical tasks have been collected, 
	the learner has the chance to learn a good prior to more reliable inferences.
	To get a good prior, it is helpful to extract mutual knowledge from similar tasks.
	Figure \ref{fig:ill} schematically illustrates this process.
	In particular, there are $m$ sample tasks $T_1,\ldots, T_m$ i.i.d.\ generated from an unknown task distribution $\tau$.
	For each sample task $T_i$, a sample set $S_i$ with $n_i$ samples is generated from an unknown distribution $D_i$.
	Without ambiguity, we use the notation $\xi(\xi^0,S)$ to denote the posterior under the prior $\xi^0$ after observing the sample set $S$. 
	Note that, the proposed meta-learner still works if the base-learner of getting the posterior is replaced by other popular methods.
	The quality of a prior $\xi^0$ is measured by $\E_{D_i\sim\tau}\E_{S_i\sim D_i^{n_i}}R(\xi(\xi^0,S_i),D_i)$.
	Thus, the expected loss we want to minimize is $$R(\xi^0,\tau)=\E_{D_i\sim\tau}\E_{S_i\sim D_i^{n_i}}R(\xi(\xi^0,S_i),D_i).$$

	The above expected risk cannot be computed directly for the unknown distribution $D_i$, thus the following empirical risk is used to estimate it:
	$$\hat{R}(\xi^0,S_1,\ldots,S_m)= \frac{1}{m}\sum_{i=1}^m \hat{R}(\xi(\xi^0,S_i^{train}),S_i^{validation}),$$
	where each sample set $S_i$ is divided into a training set $S_i^{train}$ and a validation set $S_i^{validation}$.
	
	Consider the regression setting for task $T$. Suppose the true model is
	$$y_T=f_{T}(x_T)+\sigma_{T}(x_T)\cdot\varepsilon_T,$$
	where $f_T\colon {\mathbb{R}}^d \rightarrow \bR$ is the function to be learned,
	the error term $\varepsilon_T$ is assumed to be independent of $x_T$ and has a known probability density $q(t), t\in R$ with mean $0$ and a finite variance. The unknown function $\sigma_T(x_T)$ controls the variance of the error at $X=x_T$.
	There are $n_T$ i.i.d.\ samples $\{(x_{T,i},y_{T,i})\}_{i=1}^{n_T}$ drawn from an unknown joint distribution of $(x_T,y_T)$.
	Assume that there is a candidate model set $\cH$, each of which is a function mapping feature (feature vector) to response, i.e., $h\in\cH\colon \bR^d\rightarrow\bR$.
	To take the information of the old tasks, which can reflect the importance of each $h\in\cH$, the following Algorithm \ref{alg:MAA} (meta-learner) is proposed.

	\begin{algorithm}[h]
		\caption{Historical Data Related Algorithm}
		\label{alg:MAA}
		\begin{algorithmic}[1]

			\FOR{$i=1$ to $m$}
			\STATE Using $T_i$ to obtain $\xi_i$ by the same Bayesian procedure in base-learner with non-informative prior.
			\ENDFOR
			\FOR{$i=1$ to $m$}
			\STATE Randomly split the data $S_i$ into two parts $S_{i,n_i^{'}}^{(1)}=(x_{i,\alpha},y_{i,\alpha})_{\alpha=1}^{n_i^{'}}$ for training and $S_{i,n_i^{'}}^{(2)}=(x_{i,\alpha},y_{i,\alpha})_{\alpha=n_i^{'}+1}^{n_i}$ for validation.
			\FOR{each $j\neq i$}
			\STATE Obtain estimates $\hat f_{j,n_i^{'}}(x,S_{i,n_i^{'}}^{(1)})$, $\hat\sigma_{j,n_i^{'}}(x,S_{i,n_i^{'}}^{(1)})$ with prior $\xi_j$.
			
			\STATE Evaluate predictions on $S_{i,n_i^{'}}^{(2)}$ and compute
			$$E_j^i=\frac{\Pi_{\alpha=n_i^{'}+1}^{n_i} q\left(\frac{y_\alpha-\hat f_{j,n_i^{'}}(x_{i,\alpha})}{\hat\sigma_{j,n_i^{'}}(x_{i,\alpha})}\right)}{\Pi_{\alpha=n_i^{'}+1}^{n_i}\hat\sigma_{j,n_i^{'}}(x_{i,\alpha})}.$$
			\ENDFOR
			\ENDFOR
			
			\STATE Repeat the random data segmentation more times and average the weights $E_j^i$ after normalization to get $w_j^{(i)}(j\neq i)$.
			
			\STATE Average all the $w_j^{(i)}(j\neq i)$ from $i=1$ to $m$ to obtain the final weights $w_j$.
			
			\STATE The prior learned for a new task is $\xi^*=\sum_{i=1}^mw_i\xi_i$.
			
		\end{algorithmic}
	\end{algorithm}
	
	This algorithm is based on the cross-validation framework.
	First, using $T_i$ to obtain the candidate priors $\xi_i$ by any Bayesian procedure, for example, minimizing the PAC bound introduced in Lemma \ref{thm:single} with non-informative prior.
	Cross-validation determines the importance of the priors. The $j$-th task is divided into two parts randomly. The first part is used to learn the posterior with the prior $\xi_i$. The second part is to evaluate the performance of the posterior by its likelihood function. This evaluation is inspired by \citet{Raftery1995Bayesian}.
	To simplify the determination of the weights, \citet{Raftery1995Bayesian} proposed a frequentist approach to BMA. The Bayes' theorem was replaced by the Schwarz asymptotic approximation which could be viewed as using maximized likelihood function as the weights of the candidate models.
	The $\hat\sigma$ on the denominator of $E_j^i$ makes the weight larger if the model is accurate. This procedure repeats many times for each pair $(i,j)$. Their averages reveal the importance of the priors. In the end, the $\xi^*$ is obtained by weighted averaging them all.
	The property of this algorithm can be guaranteed by Lemma \ref{thm:multi}.

	The following regularization conditions are assumed for the results.
	First, $q$ is assumed to be a known distribution with $0$ and variance $1$.
	
	\begin{itemize}
		\item[(C1)] The functions $f$ and $\sigma$ are uniformly bounded, i.e., $\sup_x|f(x)|\leq A<\infty$ and $0<m\leq\sigma(x)\leq M<\infty$ for constants $A,m$ and $M$.
		\item[(C2)] The error distribution $q$ satisfies that for each $0<s_0<1$ and $t_0>0$, there exists a constant $B$ such that
		$$\int q(x)\ln\frac{q(x)}{\frac{1}{s}q(\frac{x-t}{s})}\mu(dx)\leq B((1-s)^2+t^2)$$
		for all $s_0\leq s\leq s_0^{-1}$ and $-t_0\leq t\leq t_0$.
		\item[(C3)] The risks of the estimators for approximating $f$ and $\sigma^2$ decrease as the sample size increases.
	\end{itemize}
	
	For the condition (C1), note that when we deal with $k$-way classification tasks, the responses belong to $\{1,2,\ldots,k\}$ which is bounded obviously. Moreover, if the input space is a finite region that often happens in real datasets, the most common functions are bounded uniformly. The constants $A,m,M$ are involved in the derivation of the risk bounds, but they can be unknown in practice when we implement the Algorithm \ref{alg:MAA}.
	The condition (C2) is satisfied by Gaussian, $t$ (with the degree of freedom larger than two), double-exponential, and so on. The condition (C3) usually holds for a good estimating procedure, like consistent estimators. An estimator is called consistent if the expected risk tends to zero when the experimental size tends to infinity. 
	\begin{lemma}\label{thm:multi}
		Assume three regularization conditions are satisfied. The loss function $L(h,x,y)=|y-h(x)|^2$ and $\sigma_{T_i}$ is known. Then, the combined prior $\xi^*$ as given above satisfies
		\begin{align*}
		&R(\xi^*,\tau)\leq  \inf_j\Bigg( \frac{C_1}{\sum_{i\neq j}(n_i-n_i^{'})}\\
		&+\frac{C_2}{\sum_{i\neq j}(n_i-n_i^{'})}\sum_{i\neq j}(n_i-n_i^{'}){R}(\xi_{j},D_i)\Bigg),
		\end{align*}
		with probability at least $1-\delta$, where the constant $C_1,C_2$ depend on the regularization conditions.
		
	\end{lemma}

	Note that we assume a known $\sigma_{T_i}$ just for simplifying the expression. It has a more general version for unknown $\sigma_{T_i}$.
	The proof is given briefly with unknown $\sigma_{T_i}$ in Supplementary Materials.
	
	In this general prove, it can be seen that (i) Variance estimation is also important for the Algorithm \ref{alg:MAA}. Even if a procedure estimates $f_T$ very well, a bad estimator of $\sigma_T$ can substantially reduce its weight in the final estimator.
	(ii) Under the condition (C3), the risk of a good procedure for estimating $f_T$ and $\sigma_T$ usually decreases as the sample size increases. The influence of the number of testing points $n'_i$ is quite clear. Smaller $n'_i$ decreases the first penalty term but increases the main terms that involve the risks of each $j$.
	(iii) Lemma \ref{thm:multi} reveals the vital property that if one alternative model is consistent, the combined model will also have consistency.
	
	If Algorithm \ref{alg:MAA} is used to obtain the prior from multi-tasks, we could get the following theorem theoretically by simply combining Lemmas \ref{thm:single} and \ref{thm:multi}. See supplementary materials for the detailed proofs.
	
	\begin{theorem}\label{thm}
		Assume conditions (C1), (C2) and (C3) are satisfied. The loss function $L(h,x,y)=|y-h(x)|^2$ and $\sigma_{T_i}$ is known. Then, the combined posterior $\xi^*$ as given above satisfies
		\begin{align*}
		&R(\xi^*,\tau)\leq \inf_j\left( \frac{ C_1}{\sum_{i\neq j}(n_i-n_i^{'})}\right.\\
		&+\frac{C_2}{\sum_{i\neq j}(n_i-n_i^{'})}\sum_{i\neq j}(n_i-n_i^{'})\Bigg[\hat{R}(\xi_j^*,S_{i,n_i^{'}}^{(2)})\\
		&\left.+\sqrt{\frac{\KL(\bm{\omega_j^*}||{\bm\omega_j})+\sum_{k=1}^K\omega_{j,k} \KL(Q_{j,k}^*||Q_{j,k})+\ln\frac{n_i}{\delta}}{2(n_i-1)}}\Bigg]\right)
		\end{align*}
		with probability at least $1-\delta$, where the constant $C_1,C_2$ depend on the regularization conditions, $\pi$ is the initial prior which should be non-informative prior and $\xi_j^*$ is the minimizer of Eq.(\ref{eq:single}) with $\xi^0=\xi_j$ and $S=S_{i,n_i^{'}}^{(1)}$.
		
	\end{theorem}
	
	The major contribution of Theorem \ref{thm} is that it implies a simple consequence of consistency. The penalty for adaptation (the first term in RHS of Theorem \ref{thm}) is basically of order $1/n$, which is negligible for nonparametric rates. Thus, for any bounded regression function, the combined model performs asymptotically as well as any model in the candidate model set $\cH$. The detailed expression refers to Corollary 1 in \citet{yang2000combining}. Further, even if the underlying true model is not in the candidate set, the combined model may still be able to approach the true model, e.g, there exists a sequence of models in the set approaching an optimal one.
	
	For classification tasks, the one-hot response is used which means the $i$-th response $y_i$ in the model is a vector describing the probability of each class. The loss function still uses $\sum_{i=1}^n ||y_i-\hat y_i||^2$ to maintain consistency with the regression case. The prediction of a new observation $x^*$ is to choose the class with the largest probability in $\hat y^*$. Thus, we do not need to change the condition in Theorem \ref{thm}, but just change the response to be one-hot for classification. Consequently, the results for regression still hold for classification.
	
	Besides the $l_2$ risk that we consider, other performance measures (e.g., cross entropy loss which is used in our last experiment with the MNIST dataset or hinge loss) are useful from both theoretical and practical points of view. It is thus of interest to investigate whether similar adaptation procedures exist for other loss functions and if not, what prices are one needs to pay for adaptation, which we leave for future work.

	\section{Adjusting the Prior in Base-Learner}\label{sec:base}
	
	In this section, we will discuss how to adjust the prior of models if there is no information from extra similar tasks.

	In the following, we consider an iterative procedure of adjusting the prior in the base-learner.
	In each round, the learner can sample the data according to the prior distribution in the current round.
	Such iterative procedure updates the prior step by step. Ultimately, compared with dealing with the whole data at once, this procedure of adjusting prior leads to a smaller upper bound.
	Moreover, it also gives an opportunity to choose some good sample sets for reducing the volatility of the estimators which is measured by $v(\xi,D)=\E_x\E_{h}(h(x)-\E_{h}h(x))^2$.
	The function $\hat v(\xi,B)=\frac{1}{|B|}\sum_{x\in B}\E_{h}(h(x)-\E_{h}h(x))^2$ is defined to measure the volatility of the posterior $\xi$ at the sample set $B$.
	The complete algorithm for sequential batch sampling is shown in Algorithm~\ref{alg:SBS}.

	\begin{algorithm}[t]
		\caption{Sequential Batch Sampling Algorithm}
		\label{alg:SBS}
		\begin{algorithmic}[1]
			
			\STATE Obtain a sample set $B_1$ from the sample space $\cX\times\cY$ by an initial space-filling design.
			
			\STATE Get the posterior $\xi_1$ based on the sample set $B_1$ by minimizing the risk bound with non-informative prior.
			
			\FOR{$i=2$ to $b$}
			\STATE Search next sample set $B_i$ $(|B_i|=n_b)$ with the large volatility under the current posterior $\xi_{i-1}$, i.e., $\hat v(\xi_{i-1},B_i)>\gamma_i$ where $\bm{\gamma}$ is a given constant vector.
			
			\STATE Get the posterior $\xi_i$ based on the sample set $B_i$ by minimizing the risk bound with the prior $\xi_{i-1}$.
			\ENDFOR
			
			\STATE The final posterior is $\xi_b$.
		\end{algorithmic}
	\end{algorithm}
	
\begin{figure}[t]
		\begin{minipage}{0.48\linewidth}
			\centerline{\includegraphics[width=3.0cm]{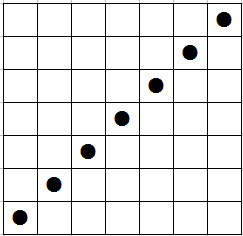}}
			\centerline{(a) Nonuniform design}
		\end{minipage}
		\hfill
		\begin{minipage}{.38\linewidth}
			\centerline{\includegraphics[width=3.0cm]{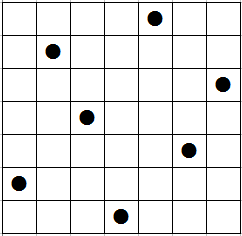}}
			\centerline{(b) Uniform design}
		\end{minipage}
		\hfill
		\caption{The illustration for uniform space-filling design.}
		\label{fig:sfd}
	\end{figure}

	For Algorithm~\ref{alg:SBS}, we do not handle the whole data at once. Instead, the data is processed in $b$ steps. First, a space-filling design is used as the initial experiment points to reduce the probability of overfitting caused by unbalanced sampling. The traditional space-filling design aims to fill the input space with design points that are as ``uniform'' as possible in the input space. The uniformity of space-filling design is illustrated in Figure \ref{fig:sfd}. For the next steps, uncertain points are needed to be explored. And, the uncertainty is measured by the volatility $v$. Hence, the batch with large volatility will be chosen. Note that if we set a huge $\gamma$, we will just explore a small region of the input space.
	
	The setting of $\bm\gamma$ refers to \cite{zhou2018sequential}.
	However, in practice, it is found that this parameter $\bm\gamma$ does not matter much since the results are similar to a wide range of $\bm\gamma$. This procedure helps to reduce the variance of the estimator which is proved in \cite{zhou2018sequential} by sequential sampling. Furthermore, it also helps to adjust the prior in each step which is called learning the prior. The proposition is stated below.
	
	\begin{prop}\label{prop:seq}
		For $i=1,2,\ldots,b$, let $B_i=S$, $\xi^*$ is the minimizer of the RHS of Eq.(\ref{eq:single}) with non-informative prior $\xi^0$ and $\xi_i$ obtained by Algorithm \ref{alg:SBS}, then we have $\overline R(\xi_b,\xi_{b-1},S)\leq\overline R(\xi^*,\xi^0,S)$.
		
	\end{prop}

	The above proposition can be understood straightforwardly. First, since we adjust the prior through the data step by step, the final prior $\xi_{b-1}$ is better than the non-informative prior. Consequently, it receives a smaller expected risk. Second, we choose the sample sets sequentially with large volatility to do experiments in order to reduce uncertainty. The performance of the Sequential Batch Sampling (SBS) method is also demonstrated in Section \ref{sec:sim}.

	\section{Experiments}\label{sec:sim}
	In this section, some examples are shown to illustrate the procedure of Algorithms \ref{alg:MAA} and \ref{alg:SBS}. The method of minimizing the upper bound in Lemma \ref{thm:single} with non-informative prior is denoted by RBM (Risk Bound Method). First, we begin with linear regression models that have the same setting in \citet{Ye2016Sparsity}. Hence, their method called SOIL is under comparison. The optimization for RHS of Eq.(\ref{eq:single}) in our algorithms is dealt with gradient descend. R package ``SOIL'' is used to obtain the results of the SOIL method. 
	
	\subsection{Synthetic Dataset}
	
	\begin{example}\label{ex:linear}
		The simulation data $\{(\bm{x_i},y_i)\}_{i=1}^n$ is generated for the RBM from the linear model $y_i=1+\bm{x_i}^T\bm\beta+\sigma\varepsilon_i$, where $\varepsilon_i\sim N(0,1)$, $\sigma \in \{1,5\}$ and $\bm{x_i}\sim N_d(0,\Sigma)$. For each element $\Sigma_{ij}$ of $\Sigma$, $\Sigma_{ij}=\rho^{|i-j|}~(i\neq j)$ or $1~(i=j)$ with $\rho\in\{0,0.9\}$.
		The sequential batch sampling has $b$ steps, and each step uses $n/b$ samples following Algorithm \ref{alg:SBS}.
	\end{example}
	
	All the specific settings for parameters are summarized in Table \ref{tab:linear}, and the confidence level $\delta$ in Lemma \ref{thm:single} is set to $0.01$. The Mean Squared Prediction Error (MSPE) $\E_x|f(x)-\hat f(x)|^2$ and volatility defined in the base-learner are compared. They are obtained by sampling $1000$ samples from the same distribution and computing their empirical MSPE $\sum_x|f(x)-\hat f(x)|^2/10^3$ and volatility. For each model setting with a specific choice of the parameters $(\rho,\sigma)$, we repeat $100$ times and compute the average empirical value. The comparison among RBM, SOIL and SBS is shown in Tables \ref{tab:reslinear1}, \ref{tab:reslinear2} and \ref{tab:reslinear3}.
	
	The volatility of the SOIL method is the smallest and very close to zero. This phenomenon shows that SOIL is focused on a few models, even just one model when the volatility equals zero. Consequently, its MSPE is larger than the other two methods. SBS as a modification of RBM has similar results with RBM when $\sigma$ is small. However, when $\sigma$ is large, SBS performs much better than RBM. In this situation, the information of data is easily covered by big noises. Hence, a good prior which can provide more information is vital for this procedure.
	
	\begin{table}[!h]
		\centering
		\caption{Simulation settings of Example \ref{ex:linear}.}\label{tab:linear}
		\vskip 0.15in
		\setlength{\tabcolsep}{1mm}{
			\begin{tabular}{l l l l l}
				\hline
				Model & n & d & b & $\bm\beta$ \\\hline
				1 & 50 & 8 & 5 & $(3,1.5,0,0,2,0,0,0)^T$\\
				2 & 150 &50 & 5 &$(1,2,3,2,0.75,0,\ldots,0)^T$\\
				3 & 50 &50 & 5 &$(1,1/2,1/3,1/4,1/5,1/6,0,\ldots,0)^T$\\
				\hline
		\end{tabular}}

		\vskip -0.1in
	\end{table}
	\begin{table}[!h]
		\centering
		\caption{Comparison among RBM, SOIL and SBS for Model 1 of Example \ref{ex:linear}.}\label{tab:reslinear1}
		\vskip 0.15in
		\begin{tabular}{l l l l l l}
			\hline
			Model 1 & ($\rho,\sigma$) & (0, 1) & (0, 5) & (0.9, 1) & (0.9, 5) \\\hline
			& RBM & 2.03 & 48.23  &  3.71    &  53.83\\
			MSPE  & SOIL     & 2.13 & 53.21  &  2.17    &  53.21\\
			& SBS      & \textbf{1.71} & \textbf{14.08}  & \textbf{ 3.25}    &  \textbf{26.40}\\
			\hline
			& RBM & 1.64 & 3.47   &  1.31    & 0.49 \\
			Volatility& SOIL & 0    & 0      &  0.002   & 0\\
			& SBS      & 1.61 & 7.41   &  1.03    & 0.42\\
			
			\hline
		\end{tabular}
		\vskip -0.1in
	\end{table}

	\begin{table}[!h]
		\centering
		\caption{Comparison among RBM, SOIL and SBS for Model 2 of Example \ref{ex:linear}.}\label{tab:reslinear2}
		\vskip 0.15in
		\begin{tabular}{l l l l l l}
			\hline
			Model 2 & ($\rho,\sigma$) & (0, 1) & (0, 5) & (0.9, 1) & (0.9, 5) \\\hline
			& RBM & 1.97 & 46.26  &    1.46   &  35.97\\
			MSPE  & SOIL     & 2.01 & 50.23  &    1.96   &  49.78\\
			& SBS      & \textbf{1.93} & \textbf{38.69}  &   \textbf{1.38}   &  \textbf{12.92}\\
			\hline
			& RBM & 1.60 & 2.72   &   3.38   & 7.48 \\
			Volatility& SOIL & 0 &  0     &   0.001      & 0.01\\
			& SBS      & 1.46 & 8.67  &   3.35   &  6.74\\
			
			\hline
		\end{tabular}
		\vskip -0.1in
	\end{table}
	
	\begin{table}[!h]
		\centering
		\caption{Comparison among RBM, SOIL and SBS for Model 3 of Example \ref{ex:linear}.}\label{tab:reslinear3}
		\vskip 0.15in
		\begin{tabular}{l l l l l l}
			\hline
			Model 3 & ($\rho,\sigma$) & (0, 1) & (0, 5) & (0.9, 1) & (0.9, 5) \\\hline
			& RBM      & 1.67 & 42.06  &  1.24    &  38.51\\
			MSPE  & SOIL     & 1.99 & 49.80  &  1.93    & 47.99 \\
			& SBS      & \textbf{1.65} & \textbf{27.32}  &  \textbf{1.23}    & \textbf{29.44} \\
			\hline
			& RBM      & 0.27 & 1.54   &  0.74    & 3.39 \\
			Volatility& SOIL & 0 &   0 &  0.02   &  0.36 \\
			& SBS      & 0.29 & 0.47   &  0.77    & 4.06\\
			
			\hline
		\end{tabular}
		\vskip -0.1in
	\end{table}

	The next example considers the same comparison but in non-linear models. In the last example, the alternative models include the true model, but now the true non-linear model is approximated by many linear models.
	
	\begin{example}\label{ex:nonlinear}
		The simulation data $\{(\bm{x_i},y_i)\}_{i=1}^{50}$ is generated for the RBM from the non-linear models
		\begin{enumerate}
			\item $y_i=1+\sin(x_{i,1})+\cos(x_{i,2})+\varepsilon_i$,
			\item $y_i=1+\sin(x_{i,1}+x_{i,2})+\varepsilon_i$,
		\end{enumerate}
		where $\varepsilon_i\sim N(0,1)$, and $\bm{x_i}\sim N_8(0,I)$.
		The sequential batch sampling has $5$ steps, and each step uses $10$ samples following Algorithm \ref{alg:SBS}.
	\end{example}
	
	The results of Example \ref{ex:nonlinear} are listed in Table \ref{tab:resnonlinear}. Mostly, it is similar to the results of Example \ref{ex:linear}. The difference is that the volatility of SOIL becomes large when the model is completely non-linear. Using linear models to fit non-linear model increases the model uncertainty since none of the fitting models is correct.
	
	\begin{table}[!h]
		\centering
		\caption{Comparison among RBM, SOIL and SBS of Example \ref{ex:nonlinear}.}\label{tab:resnonlinear}
		\vskip 0.15in
		\setlength{\tabcolsep}{4.5mm}{
			\begin{tabular}{l l l l }
				\hline
				&  & Model 1 & Model 2  \\\hline
				& RBM      & 1.26 & 1.54  \\
				MSPE  & SOIL     & 1.42 & 1.80  \\
				& SBS      & \textbf{1.23} & \textbf{1.47}  \\
				\hline
				& RBM      & 0.1 & 0.11   \\
				Volatility& SOIL & 0.07 &   0.02  \\
				& SBS      & 0.11 & 0.14  \\
				
				\hline
		\end{tabular}}
		\vskip -0.1in
	\end{table}
	
	The final example is under the situation that the data has been already collected. Hence, we cannot use the SBS method to get the data. However, we have the extra data of many old similar tasks. In particular, we have the data of Example \ref{ex:linear}. Now, the new task is to fit a new model.
	
	\begin{example}\label{ex:mul}
		The data of Example \ref{ex:linear} with $(\rho,\sigma)=(0,1)$ is given. The new task data $\{(\bm{x_i},y_i)\}_{i=1}^{20}$ is generated from the linear model $y_i=1+\bm{x_i}^T\bm\beta+\sigma\varepsilon_i$, where $\varepsilon_i\sim N(0,1)$, $\sigma \in \{1,2,3,4,5\}$, $\bm\beta=\{1,-1,0,0,0.5,0,\ldots,0\}$ and $\bm{x_i}\sim N_{10}(0,I)$.
		
	\end{example}
	
\begin{figure}
		\centering
		\includegraphics[width=6.9cm]{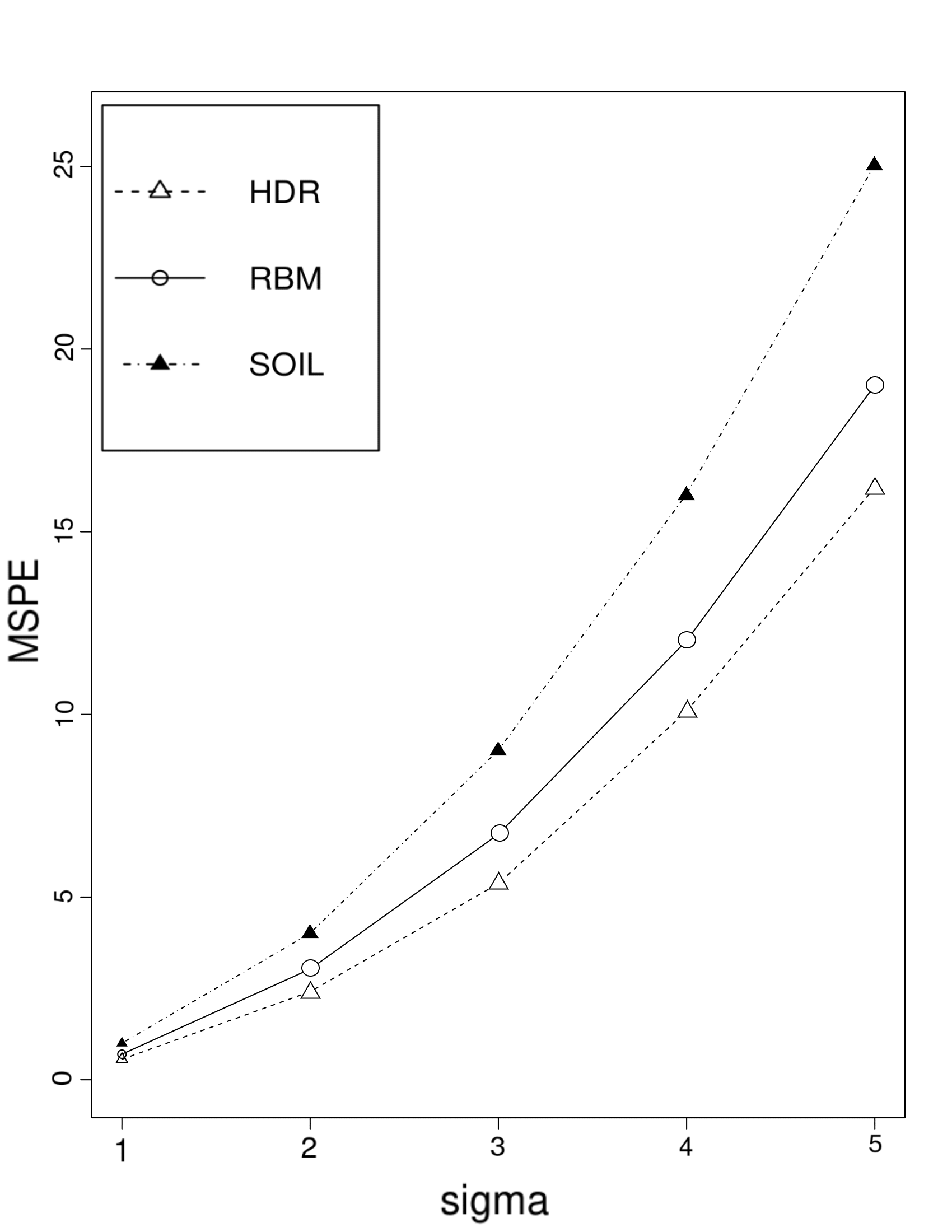}\\
		\caption{Comparison among RBM, SOIL and HDR of Example \ref{ex:mul}.}\label{fig:resmul}
	\end{figure}

	The method described in Algorithm \ref{alg:MAA} is denoted by HDR (Historical Data Related). The results in Figure \ref{fig:resmul} show high consistency with the last two examples.
	When $\sigma$ is small, the different priors lead to a similar result since the current data has a key influence. However, when $\sigma$ is large, the difference between RMB and HDR is huge.
	The reason is that the current data has been polluted by the strong noise. Hence, a good prior can provide vital information about the model distribution.
	
	\subsection{Real-World Dataset}
	
	Here, we apply the proposed methods to two real datasets, BGS data and Bardet data, which are also used in \citet{Ye2016Sparsity}.
	
	First, the BGS data is with a small $d$ and from the Berkeley Guidance Study (BGS) by \citet{tuddenham1954physical}. The dataset records $66$ boys' physical growth measures from birth to eighteen years. Following \citet{Ye2016Sparsity}, we consider the same regression model. The response is age $18$ height and the factors include weights at ages two (WT2) and nine (WT9), heights at ages two (HT2) and nine (HT9), age nine leg circumference (LG9) and age $18$ strength (ST18).
	
	Second, for large $d$, the Bardet data collect tissue samples from the eyes of $120$ twelve-week-old male rats.
	For each tissue, the RNAs of $31,042$ selected probes are measured by the normalized intensity valued. The gene intensity values are in log scale.
	Gene TRIM32, which causes the Bardet-Biedl syndrome, is the response in this study. The genes that are related to it are investigated.
	A screening method \cite{Huang2008Adaptive} is applied to the original probes. This screened data with $200$ probes for each of $120$ tissues are also used in \citet{Ye2016Sparsity}.
	
	Both cases are data-given cases that we cannot use the sequential batch sampling method. For the different settings of $d$, we assign corresponding similar historical data for two real datasets. The data of model $1$ in Example \ref{ex:linear} for the BGS data with small $d$. The data of model $3$ in Example \ref{ex:linear} for the Bardet data with large $d$.
	
	We randomly sample $10$ rows from the data as the test set to calculate empirical MSPE and volatility. The results are summarized in Table \ref{tab:real}. From Table \ref{tab:real}, we can see that both RBM and HDR have smaller MSPE than SOIL. However, HDR does not perform much better than RBM. This can be explained intuitively as follows. In theory, the historical tasks and the current task are assumed that they come from the same task distribution. But in practice, how to measure the similarity between tasks is still a problem. Hence, an unrelated historical dataset may provide less information for the current prediction.
	\begin{table}[!h]
		\centering
		\caption{Comparison among RBM, SOIL and HDR in real data.}\label{tab:real}
		\vskip 0.15in
		\setlength{\tabcolsep}{5mm}{
			\begin{tabular}{l l l l }
				\hline
				
				&  & BGS & Bardet  \\\hline
				& RBM      & 13.54 & 0.0054  \\
				MSPE  & SOIL     & 16.74 & 0.0065  \\
				& HDR      & \textbf{13.06} & \textbf{0.0050}  \\
				\hline
				& RBM      & 1.99 & 0.0013   \\
				Volatility& SOIL & 0.43 &   0.0013 \\
				& HDR      & 1.84 & 0.0012  \\
				
				\hline
		\end{tabular}}
		\vskip -0.1in
	\end{table}

	\subsection{Classification Tasks}\label{}

	In this section, the performance of our HDR method for classification tasks is demonstrated. Here, we use the same image classification example in \citet{amit2018meta-learning}. The hypothesis class is the set of neural networks including the architecture given in \citet{amit2018meta-learning} and other CNN architectures provided in Keras. Different architectures are weighted by the parameter $\bm\omega$. And, the distribution $Q_i$ is to characterize the hyperparameter of the $i$-th architecture. The cross-entropy loss is used.
	
	The task environment is constructed based on augmentations of the MNIST dataset \cite{Lecun1998}. Each task is created by a permutation of the image pixels. We randomly pick $100$, $200$ and $300$ pixel swaps, and find that they have similar results. Thus we just show the results with $200$ pixel swaps. For the meta-learner, it is trained by the meta-training tasks each with $50000$ training samples and $10000$ validation samples. For a new task with fewer training samples and $10000$ test samples, we randomly sample $2000$ training samples $20$ times and compare the average test error percentage of different learning methods. `$95\%$ CI' in Table \ref{tab:pixel} means the $95\%$ confidence interval. MLAP represents the method in \citet{amit2018meta-learning}, and different subscripts mean that different PAC bounds are used. The Model-Agnostic-Meta-Learning \cite{finn2017model} is denoted by MAML. See \citet{amit2018meta-learning} for detailed settings of the example.
	
	Table \ref{tab:pixel} summarizes the results for the permuted pixel environment with $200$ pixel swaps and $10$ training tasks. We find that the best results are obtained by our HDR method. Note that for classification tasks, the weighted prediction is for the one-hot response. Hence, the prediction can be viewed as picking the largest probability among all the models not just in one model. Consequently, model averaging used in classification has much better performance than regression cases.
	
	We also investigate whether the number of training tasks affects the error rate of the predictions on the new test tasks, and find that it is improved a lot if the meta-learner is used. But, the number of training tasks does not have a significant effect.
	
	\begin{table}[t]
		\centering
		\caption{Comparisons of different learning methods on $20$ test tasks of classification.}\label{tab:pixel}
		\vskip 0.15in
		\setlength{\tabcolsep}{7mm}{
			\begin{tabular}{l l l }
				\hline
				METHOD & ERROR & $95\%$ CI  \\\hline
				MLAP$_M$      & 3.4 & 0.18  \\
				MLAP$_S$      & 3.54 & 0.2  \\
				MLAP$_{PL}$      & 74.9 & 4.03 \\
				MLAP$_{VB}$      & 3.52 & 0.17   \\
				MAML   & 3.77 &   0.8 \\
				HDR      & \textbf{0.72}  & 0.0003  \\
				
				\hline
		\end{tabular}}
		\vskip -0.1in
	\end{table}

	\section*{Acknowledgment}
	We sincerely thank Prof. Zhihua Zhang (Peking University) and all reviewers for their valuable comments which have led to further improvement of this work.
	
	\bibliographystyle{aaai}
	\bibliography{10447-ref}

	\clearpage
	\section*{Supplementary Materials}
	For Lemma \ref{thm:single}, we review the classical PAC-Bayes bound \cite{Mcallester1999PAC} with general notations first.
	
	\begin{lemma}\label{thm:classical}
		Let $\cX$ be a sample space and $\cF$ be a function space over $\cX$. Define a loss function $g(f,X):\cF\times\cX\rightarrow[0,1]$, and $S=\{X_1,\ldots,X_n\}$ be a sequence of $n$ independent identical distributed random samples. Let $\pi$ be some prior distribution over $\cF$. For any $\delta\in(0,1]$, the following bound holds for all posterior distributions $\rho$ over $\cF$,
		\bea\label{eq:classical}
		\begin{aligned}
			\bP_S\bigg(\E_X\E_f g(f,X)&\leq\frac{1}{n}\sum_{i=1}^n\E_f g(f,X_i)\\
			&+\sqrt{\frac{\rho||\pi+\ln\frac{n}{\delta}}{2(n-1)}}\bigg)\geq1-\delta.
		\end{aligned}
		\eea
	\end{lemma}
	\textbf{Proof of Lemma \ref{thm:single}}:
	We use Lemma \ref{thm:classical} to bound the expected risk with the following substitutions. The $n$ samples are $X_i\triangleq z_i$. The function $f\triangleq h$ where $h\in\cH$.
	The loss function $g(f,X)\triangleq L(h,z)\in[0,1]$. The prior $\pi$ is defined by $\pi\triangleq \xi^0$, in which we first sample $k$ from $\{1,\ldots,K\}$ according to corresponding weights $\{w_1,\ldots,w_K\}$ and then sample $h$ from $Q_k$. The posterior is defined similarly, $\rho\triangleq\xi$.
	
	The KL-divergence term is
	
	\bea
	\begin{aligned}
		\rho||\pi&=\E_f\ln\frac{\rho(f)}{\pi(f)}=\E_{k\in\{1,\ldots,K\}}\E_{h\in\cM_k}\ln\frac{w_kQ_k(h)}{w_{0,k}Q^0_{k}(h)}\\
		&=\sum_{k=1}^Kw_k\E_{h\in\cM_k}\ln\frac{w_kQ_k(h)}{w_{0,k}Q^0_{k}(h)}\\
		&=\bm{\omega}||{\bm\omega^0}+\sum_{k=1}^K\omega_k (Q_k||Q^0_{k}).
	\end{aligned}
	\eea
	
	Substituting the above into Eq.(\ref{eq:classical}), it follows that
	
	\bea
	\begin{aligned}
		&\bP_S\Bigg(\E_z\E_{k\in\{1,\ldots,K\}}\E_{h\in\cM_k}L(h,z)\\
		&\leq\frac{1}{n}\sum_{i=1}^n\E_{k\in\{1,\ldots,K\}}\E_{h\in\cM_k}L(h,z)\\
		&+\sqrt{\frac{\bm{\omega}||{\bm\omega^0}+\sum_{k=1}^K\omega_k (Q_k||Q^0_{k})+\ln\frac{n}{\delta}}{2(n-1)}}\Bigg)\geq1-\delta.
	\end{aligned}
	\eea
	
	Using the notations in Section \ref{sec:meta}, we can rewrite the above as below,
	\bea\label{eq:single2}
	\begin{aligned}
		\bP_S\Bigg(&R(\xi,D)\leq\hat{R}(\xi,S)\\
		&+\sqrt{(\bm{\omega}||{\bm\omega^0}+\sum_{k=1}^K\omega_k (Q_k||Q^0_{k})+\ln\frac{n}{\delta})/(2n-2)}\Bigg)\\
		&\geq1-\delta.
	\end{aligned}
	\eea
	\qed
	
	\textbf{Proof of Proposition \ref{prop:seq}}:
	
	First, we prove that for $i=2,\ldots,b$, $$\overline R(\xi_i,\xi_{i-1},B_i)\leq\overline R(\xi_{i-1},\xi_{i-2},B_{i-1}).$$
	By definition of $\xi_i$,
	\begin{align*}
	&\overline R(\xi_i,\xi_{i-1},B_i)\leq\overline R(\xi_{i-1},\xi_{i-1},B_{i})\\
	&=\hat R(\xi_{i-1},B_i)+\sqrt{\ln\frac{n}{\delta}/(2n-2)}\\
	&\leq \overline R(\xi_{i-1},\xi_{i-2},B_{i})=\overline R(\xi_{i-1},\xi_{i-2},B_{i-1}).
	\end{align*}
	Following these inequalities,
	\begin{align*}
	&\overline R(\xi_b,\xi_{b-1},S)=\overline R(\xi_{b},\xi_{b-1},B_{b})\\
	&\leq\overline R(\xi_{1},\xi^{0},B_{1})=\overline R(\xi^*,\xi^{0},S).
	\end{align*}
	This finishes the proof.
	\qed
	
	\textbf{Proof of Lemma \ref{thm:multi}}:

	According to Theorem $1$ in \cite{Yang2001Adaptive}, we have
	
	\bea\label{eq:arm}
	\begin{aligned}
		&R(\xi^*,\tau)\leq \inf_j\Bigg( \frac{ C_1}{\sum_{i\neq j}(n_i-n_i^{'})}\\
		&+\frac{C_2}{\sum_{i\neq j}(n_i-n_i^{'})}\sum_{i\neq j}\sum_{\alpha=n_i^{'}+1}^{n_i}\Bigg[\E||\sigma_{T_i}^2-\hat\sigma_{j,\alpha}^2||^2\\
		&+{R}(\xi_{j}^*,D_i)\Bigg]\Bigg),
	\end{aligned}
	\eea
	where $\xi_j^*$ is the minimizer of Eq.(\ref{eq:single}) with $\xi_0=\xi_j$ and $S=S_{i,\alpha}^{(1)}$ denoted by $\xi_j^*(\xi_j,S_{i,\alpha}^{(1)})$.
	
	For any $\alpha\geq n_i^{'}$ and an estimator satisfied the condition (C3), the inequalities $\E||\sigma_{T_i}^2-\hat\sigma_{j,n_i^{'}}^2||^2\geq\E||\sigma_{T_i}^2-\hat\sigma_{j,\alpha}^2||^2$ and ${R}(\xi_{j}^*(\xi_j,S_{i,n_i^{'}}^{(1)}),D_i)\geq{R}(\xi_{}^*(\xi_j,S_{i,\alpha}^{(1)}),D_i)$ hold.
	Plugging into (\ref{eq:arm}) for $\alpha=n_i^{'}+1,\ldots,n_i$, it follows that
	
	\begin{align*}
	&R(\xi^*,\tau)\leq  \inf_j\Bigg( \frac{C_1}{\sum_{i\neq j}(n_i-n_i^{'})}\\
	&+\frac{C_2\sum_{i\neq j}(n_i-n_i^{'})\Big[\E||\sigma_{T_i}^2-\hat\sigma_{j,n_i^{'}}^2||^2+{R}(\xi_{j}^*,D_i)\Big]}{\sum_{i\neq j}(n_i-n_i^{'})}\Bigg),
	\end{align*}
	
	where $\xi_j^*$ is the minimizer of Eq.(\ref{eq:single}) with $\xi^0=\xi_j$ and $S=S_{i,n_i^{'}}^{(1)}$.
	
	In order to obtaining the form in Theorem \ref{thm:multi}, it only needs to note that if $\sigma_{T_i}$ is known, the term $\E||\sigma_{T_i}^2-\hat\sigma_{j,n_i^{'}}^2||^2$ vanishes. \qed

\end{document}